\documentclass[10pt,journal,letterpaper]{IEEEtran} 
\usepackage{amsmath}
\usepackage{amssymb}
\usepackage{mathrsfs}
\usepackage{graphicx}
\usepackage{cite}
\usepackage{multirow}
\usepackage{multicol}
\usepackage{array}
\usepackage{minipage-marginpar}
\usepackage{subfigure}
\usepackage[numbers,sort&compress]{natbib}
\usepackage{paralist}
\usepackage{hyperref}
\usepackage{todonotes}
\usepackage{color,soul}

\ifCLASSINFOpdf
\else
\fi
\begin{document}
%
\title{Real-Time Quality Assessment of Pediatric MRI via Semi-Supervised Deep Nonlocal Residual Neural Networks}

\author{Siyuan Liu,~\IEEEmembership{Member,~IEEE,} Kim-Han Thung, Weili Lin, Pew-Thian Yap,~\IEEEmembership{Senior Member,~IEEE}, Dinggang Shen,~\IEEEmembership{Fellow,~IEEE} 
	\thanks{S.~Liu, K.-H.~Thung, W.~Lin, P.-T.~Yap and D.~Shen are with the Department of Radiology and Biomedical Research Imaging Center (BRIC), University of North Carolina at Chapel Hill, NC, U.S.A. D.~Shen is also with the Department of Brain and Cognitive Engineering, Korea University, Seoul, Korea.}
  \thanks{Corresponding authors: ptyap@med.unc.edu, dgshen@med.unc.edu}
}


%


\maketitle

\begin{abstract}
In this paper, we introduce an image quality assessment (IQA) method for pediatric T1- and T2-weighted MR images. IQA is first performed slice-wise using a nonlocal residual neural network (NR-Net) and then volume-wise by agglomerating the slice QA results using random forest. Our method requires only a small amount of quality-annotated images for training and is designed to be robust to annotation noise that might occur due to rater errors and the inevitable mix of good and bad slices in an image volume. Using a small set of quality-assessed images, we pre-train NR-Net to annotate each image slice with an initial quality rating (i.e., pass, questionable, fail), which we then refine by semi-supervised learning and iterative self-training. Experimental results demonstrate that our method, trained using only samples of modest size, exhibit great generalizability, capable of real-time (milliseconds per volume) large-scale IQA with near-perfect accuracy.
\end{abstract}




%
\IEEEpeerreviewmaketitle

\section{Introduction}
Structural magnetic resonance imaging (sMRI) is widely used for brain morphological analysis due to its high spatial-resolution details of anatomical structures.
However, sMRI is susceptible to image artifacts caused for instance by eye and head motion, hemodynamic changes, and magnetic field inhomogeneities \cite{Zhuo2006MR}. 
Among these artifacts, motion artifacts are particularly prevalent when scanning pediatric subjects. 
As poor-quality images may bias subsequent analysis and result in incorrect conclusions, it is vital to correctly identify problematic images and exclude them from analysis.

Image quality assessment (IQA) is an important step to determine whether the acquired data are usable and whether a re-scan is necessary. 
IQA can be performed subjectively by a human rater or objectively by a computer algorithm.
The most commonly used subjective quality ratings can be grouped into two categories, i.e.,
\begin{inparaenum}
	\item score-based rating, where a visual quality score metric, such as Mean Opinion Score (MOS) \cite{Huynh-Thu2011Study}, is used for quality grading; and
	\item class-based rating, where a visual quality class spectrum, such as Excellent/Very Good/Good/Fair/Poor/Unusable \cite{Kustner2017An} or Pass (Excellent to Very Good)/Questionable (Good to Fair)/Fail (Poor to Unusable) \cite{White2017Automated}, is used for quality grading.
\end{inparaenum}
Subjective IQA with visual inspection, even when carried out by experienced radiologists, is time-consuming, labor-intensive, costly, and error-prone \cite{Elias2008Automated}. 
Therefore, a reliable, accurate, and fully-automated objective IQA of sMRI is highly desirable.

Based on the availability of a reference image, objective IQA can be grouped into three categories, i.e.,
\begin{inparaenum}[(i)]
	\item full-reference IQA (FR-IQA), which requires a pristine image as reference; 
	\item reduced-reference IQA (RR-IQA), which requires partial information from a reference image; and 
	\item no-reference IQA (NR-IQA) or blind IQA, which requires no reference image. 
\end{inparaenum}
FR-IQA measures the quality of an image by comparing it with a reference using some evaluation metrics.
On the other hand, RR-IQA uses only a limited number of features extracted from a reference \cite{Rehman2012Reduced} to provide a near FR-IQA performance.
However, FR-IQA and RR-IQA have limited practical application as full or partial information from a pristine reference image is not always available.
For quality assessment without any reference image, as in our case, NR-IQA \cite{Manap2015Non} is needed. Currently, most NR-IQA methods \cite{Ye2012No,Tang2011Learning,Saad2012Blind} are designed for natural 2D images. MR images are typically 3D with intensity distributions and artifacts that are very different from natural images.

Recently, deep neural networks (DNNs), particularly convolutional neural networks (CNNs), have demonstrated great potential for IQA \cite{Kim2017Deep}.
Instead of hand-crafted features, CNNs automatically learn image features that are pertinent to IQA. 
However, the performance of these deep learning methods generally depends on a large number of correctly labeled training samples, which are typically lacking for medical images, as labeling involves a huge amount of effort from experts. For example, it is labor-intensive and time-consuming to annotate quality scores for all the image slices in MR images to train a slice-wise IQA network. Thus, annotation is typically performed volume-wise, where each MR volume is associated with a single quality label. This is however inaccurate since each volume might contain a mix of good or bad slices. 

In this paper, we address the above issues by introducing a deep learning based slice- and volume-wise IQA method that is robust to annotation errors and requires only a small amount of annotated images for training. To the best of our knowledge, this is the first work on deep learning based slice- and volume-wise NR-IQA of sMRI with explicit consideration of limited training samples and labeling noise. 
The key features of our method are summarized as follows:
\begin{enumerate}
\item 
Our method consists of a nonlocal residual neural network (NR-Net) for slice-wise IQA and a random forest to agglomerate the slice IQA results for volume-wise IQA. 
We train our NR-Net using the slices of annotated image volumes, effectively increasing training sample size.

\item We employ depthwise separable residual (DSRes) blocks \cite{Chollet2017Xception} and nonlocal residual (NRes) blocks \cite{Wang2018Non-local} to construct the NR-Net. 
Compared to residual networks with standard convolutions, the computation-reduction property of the DSRes block and also the information fusion property of the NRes block make NR-Net much lighter, hence allowing fast real-time IQA.

\item We utilize semi-supervised learning to deal with the scenario where we have a small amount of labeled data but a large amount of unlabeled data. With a small number of labeled samples, we pre-train the NR-Net to label the slices of unlabeled volumes, which are then used to re-train the NR-Net.

\item We use an iterative self-training mechanism to prune or relabel unreliable labels to improve training effectiveness. Self-training is iterated until convergence.

\end{enumerate}

The rest of this paper is organized as follows. 
In Section~\ref{sec:RelatedWork}, we briefly review related work on NR-IQA. 
Section~\ref{sec:Architecture} describes the NR-Net architecture, semi-supervised learning, and slice/volume self-training. 
We present the experimental results in Section~\ref{sec:Experiments} and conclude this paper in Section~\ref{sec:Conclusions}.

\section{Related Work}\label{sec:RelatedWork}

\subsection{Deep Learning NR-IQA}
Deep learning NR-IQA methods can be either score-based or class-based \cite{Yang2019A}.
Score-based methods predict continuous-valued quality ratings, typically by treating quality assessment as a regression problem.
Deep learning NR-IQA methods in this category (mostly for 2D natural images) \cite{Liu2017Rank,Lin2018Hallucinated,Ma2018End,Kim2018Multiple,Zhang2020Blind} are typically trained image-wise and thus a large number of annotated images are needed. Patch-wise methods \cite{Guan2017Visual,Bosse2018Deep,Yan2019Two,Kim2019Deep} use image patches as training samples, thus effectively increasing sample size.

Class-based methods \cite{Hou2015Blind,Hou2015Saliency,Li2015No,Pizarro2016Automated,Bianco2017On,Esses2017Automated,Gu2018Blind,Muelly2017Automated,Kuestner2017Automated,Ding2018Supervised} predict discrete-valued quality categories, often associated with human perception on image quality. 
Quality annotation in structural MRI is carried out manually via visual inspection. There is no reliable metric that can be used to provide continuous-valued quality ratings. Therefore, our IQA method is based on classification instead of regression to predict quality categories of slices and volumes.

\subsection{Label Noise}

Expert annotations are not always accurate, thus leading to ``label noise''.
Methods to cope with label noise can be divided into three categories \cite{Frenay2014Classification}: 
\begin{inparaenum}
	\item designing a robust loss, 
	\item cleansing noisy data, and 
	\item modeling the label noise distribution.
\end{inparaenum}

First, loss function can be reweighted or rectified  to be robust to label noise \cite{Reed2014Training, Li2017Learning, Patrini2017Making}. 

Second, label noise can be ``cleaned'' before training, by either relabeling or removing mislabeled data \cite{Sebastien2007Automatic, Andre2009Use}.
Removing mislabeled data, while shown to be effective \cite{Sebastien2007Automatic, Andre2009Use}, reduces sample size and hence degrade training effectiveness.
Data cleansing is also affected by data imbalance \cite{Seiffert2014An}, where minority classes may more likely be entirely removed. In comparison, relabeling mislabeled data maintains the sample size, but incorrect relabeling may lead to performance degradation.  

Third, if some information about the label noise is available, it is possible to predict the distribution of label noise and use it to improve the classifier. Explicitly modeling or learning the label noise distribution \cite{Kaster2011Comparative, Xu2005Survey, Mnih2012Learning, Miao2016RBoost} allows noisy labels to be detected and discarded during training.  
However, this approach depends on the accuracy of the label noise model, increases the complexity of learning algorithms, and may result in overfitting due to additional model parameters.

\section{Architecture}\label{sec:Architecture}
\begin{figure*}[!t]
	\centering
	\includegraphics[width=0.98\textwidth]{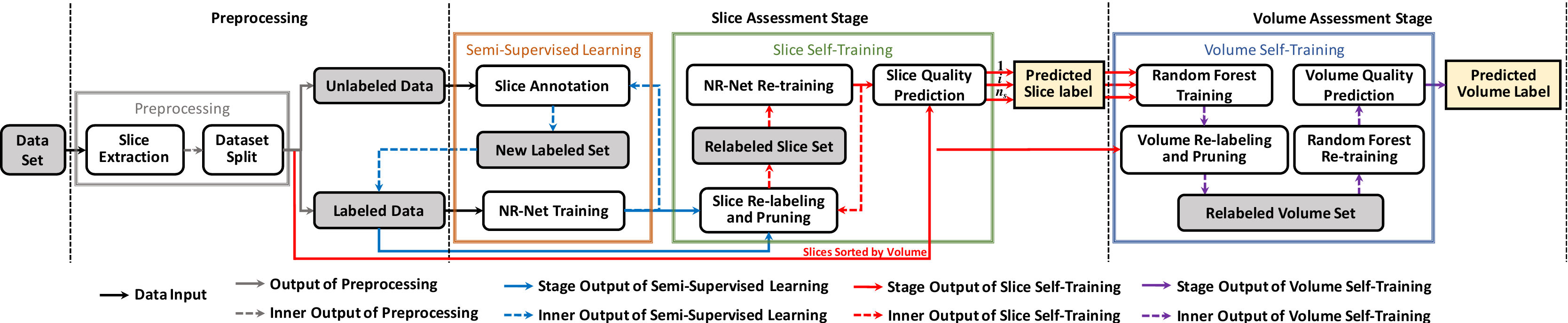}\\
	\caption{Overview of our sMRI IQA method.}
	\label{Main Flow}
\end{figure*}

Fig. \ref{Main Flow} shows an overview of our sMRI IQA method, which consists of two stages, i.e., slice assessment stage and volume assessment stage. 
The slice assessment stage is designed to predict the quality rating of each slice and is trained using semi-supervised learning and slice self-training. The volume assessment stage, trained using volume self-training, evaluates the quality rating of each volume by ensembling the quality ratings of slices belonging to this volume. 
Details of our method are described next.

\subsection{Slice Quality Assessment Network}
\begin{figure}[!t]
	\centering
	\includegraphics[width=8.5cm]{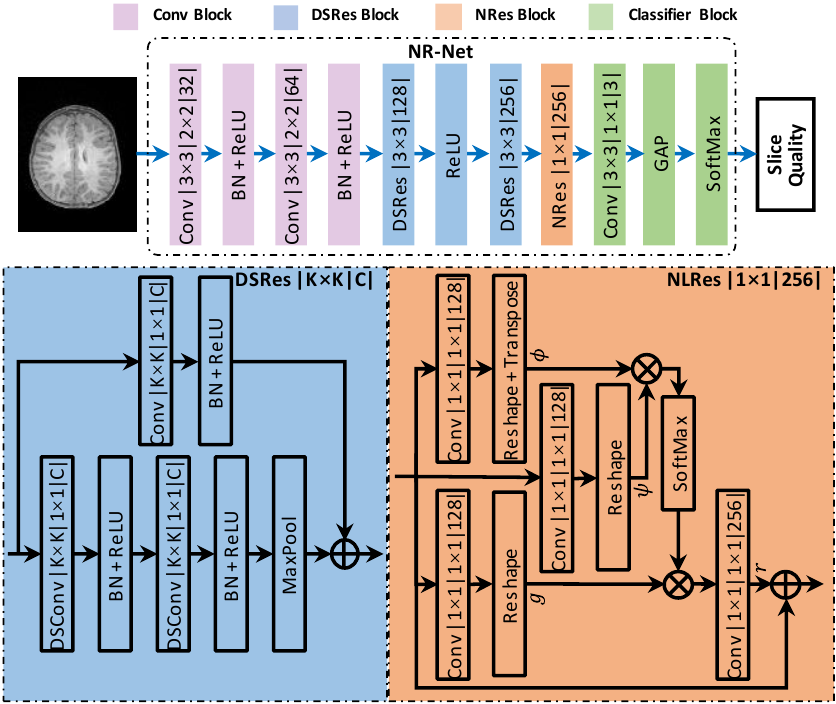}\\
	\caption{Architecture of NR-Net, which consists of four types network blocks, i.e., the convolution (Conv), depthwise separable residual (DSRes), nonlocal residual (NRes), and classifier blocks. The parameters of DSRes and NRes blocks are denoted in the following format: ``DSRes/NRes $|$ kernel size $|$ output channel $|$'', and the parameters of the convolution and depthwise separable convolution layers are similarly denoted as ``Conv/DSConv $|$ kernel size $|$ strides $|$ output channel $|$''. $\otimes$: matrix multiplication, $\oplus$: element-wise summation.}
	\label{Network Architecture}
\end{figure}

Our slice quality assessment network is designed with both accuracy and speed in mind.
Fig. \ref{Network Architecture} shows the proposed network, NR-Net, which consists of four types of network blocks, i.e., convolution (Conv), depthwise separable residual (DSRes), nonlocal residual (NRes), and classifier blocks. 
The Conv and DSRes blocks extract low- and high-level features, respectively, whereas 
the NRes block \cite{Wang2018Non-local} computes the response function at each position as a weighted summation of features from different spatial locations.  
The classifier block (realized with a convolutional layer, global average pooling, and a softmax activation function) outputs three probability values indicating whether a slice is ``pass", ``questionable", or ``fail". 
The slice is finally annotated with the label associated with the highest probability.


\begin{figure*}[!t]
  \centering
  \subfigure[Standard Convolution]{
    \includegraphics[width=0.30\textwidth]{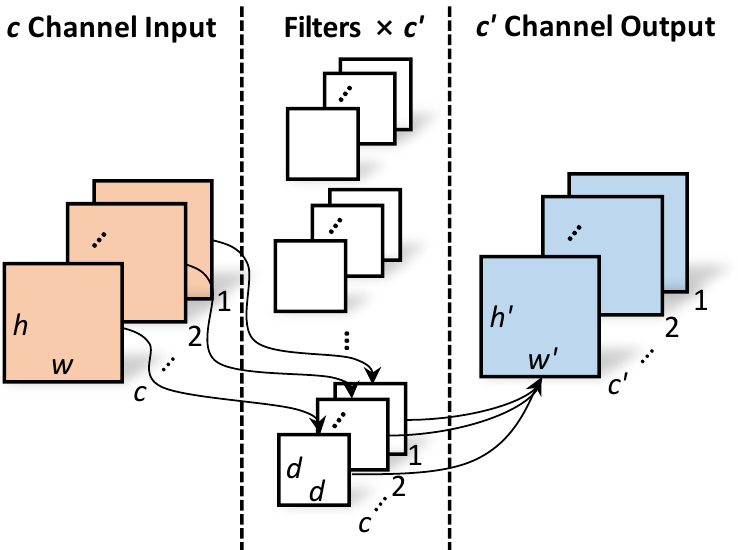}\label{fig:Conv}}
  \centering
  \subfigure[Depthwise Convolution]{
    \includegraphics[width=0.30\textwidth]{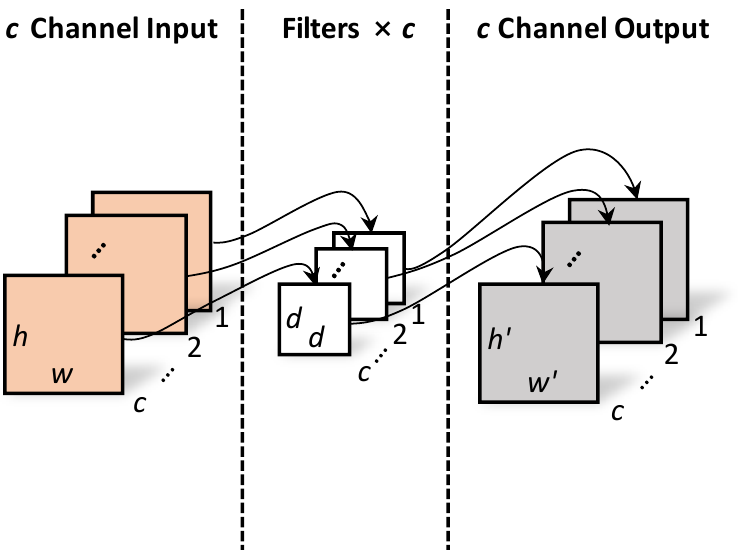}\label{fig:DepConv}}
  \centering
  \subfigure[Pointwise Convolution]{
    \includegraphics[width=0.30\textwidth]{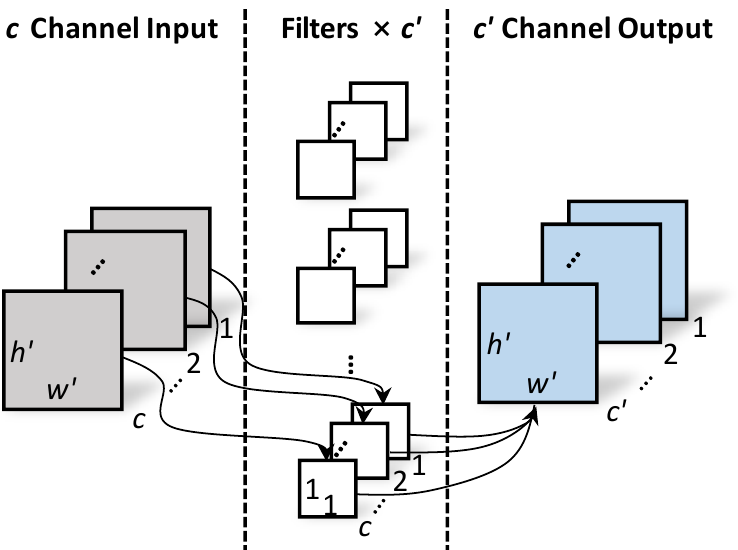}\label{fig:PointConv}}
  \caption{Differences between (a) standard convolution (Conv) and depthwise separable convolution (DSConv), which consists of (b) depthwise convolution and (c) pointwise convolution.}
  \label{fig:DSConv}
\end{figure*}

\subsubsection{DSRes Block}
We construct the DSRes block by integrating depthwise separable convolution (DSConv) layers in the NR-Net (blue in Fig.~\ref{Network Architecture}). 
Fig.~\ref{fig:Conv} shows how a standard convolution layer filters all input channels and combine the results in an output channel. 
In comparison, a DSConv layer filters using a combination of depthwise convolution and pointwise convolution. Specifically, depthwise convolution performs channel-wise spatial convolution and concatenation, as shown in Fig.~\ref{fig:DepConv}. 
Pointwise convolution subsequently projects the channels by depthwise $1\times1$ convolution onto a new channel, as shown in Fig.~\ref{fig:PointConv}.

Based on Fig.~\ref{fig:DSConv}, we show that DSConv is computationally more efficient than standard convolution.  Given a $c$-channel $h\times w$ input feature map, a $c'$-channel $h' \times w'$ output feature map, and a $d\times d$ kernel,
the computational cost (CC) of a standard convolution layer is given by $\textrm{CC}_\textrm{StdConv} = c c' d^2 h' w'$.
In contrast, the computational cost of a DSConv layer with the same input and output, as shown in Fig. \ref{fig:DSConv}(b) and (c), is given by $\textrm{CC}_\textrm{DSConv} = c d^2 h' w' + c c' h' w'$. 
The computational reduction factor (CRF) of replacing standard convolution with DSConv is therefore given as $\textrm{CRF}_\textrm{conv} = 1/c'+1/{d^2}$.
Furthermore, the CRF of replacing a residual block constructed by standard convolutions with a DSRes block is $\textrm{CRF}_\textrm{DSRes}\approx1/c'+3/(2d^2+1)$ with $c'=2c$. 
We use a $3\times 3$ convolution kernel ($d=3$) in each DSRes block, so that the computational cost of a DSRes block is 6 to 7 times smaller than that of a residual block with standard convolution. 

\subsubsection{NRes Block}
The quality of an image slice is determined by the existence of artifacts at different spatial locations. 
We employ a nonlocal residual (NRes) block \cite{Wang2018Non-local} in our network (orange in Fig.~\ref{Network Architecture}) to capture information from potentially distant locations. 
Given a $c$-channel $h\times w$ input feature map $x$, the output at the $i$-th location, $r_{i}\in\mathbb{R}^{c}$, of the NRes block is computed as the weighted sum of all features:
\begin{equation}
r_{i} = \frac{1}{C(x_{i})}\sum\limits_{\forall j}f(x_{i}, x_{j})g(x_{j}),~x_{i},x_{j}\in\mathbb{R}^{c}
\end{equation}
where the weight function $f(\cdot,\cdot)$ encodes the pairwise similarity between feature vectors at locations $i$ and $j$, $g(\cdot)$ computes a representation of a feature vector, and $C(x_{i})=\sum_{\forall j}f(x_{i}, x_{j})$ is a normalization factor.
In this work, the weight function $f(x_{i},x_{j})$ is defined as
\begin{equation}
f(x_{i}, x_{j}) = \exp\left[{\phi^T(x_{i})\psi(x_{j})}\right],
\end{equation}
where $\phi(\cdot)$ and $\psi(\cdot)$ are unary kernel functions that are implemented with $1\times 1$ convolution kernels, thus making $\frac{1}{C(x_{i})}f(x_{i}, x_{j})$ a softmax function. 
The NRes block is incorporated in the NR-Net using a residual form.

NR-Net is lighter with less parameters by employing DSRes blocks and captures long-range dependencies between features regardless of their positional distances. Batch normalization and global average pooling are used for regularization without dropout to speed up training.

\subsection{Semi-Supervised Learning}
We employ semi-supervised learning to make full use of a small amount of labeled data and a large amount of unlabeled data. This is done by progressive annotation of unlabeled slices to retrain the network (orange box in Fig.~\ref{Main Flow}). 
We begin by utilizing the NR-Net pre-trained with the labeled dataset to predict the ``pass'', ``questionable'', or ``fail'' probabilities of the slices of the unlabeled volumes. Each slice is annotated with the quality rating associated with the maximal probability.
The labeled slices are then merged into the original labeled dataset to be used for retraining of the NR-Net.

\subsection{Slice Self-Training}
To deal with noisy labels, we propose a slice self-training method to sample ``clean'' data for training.
This involves iterative slice relabeling/pruning and NR-Net retraining (green box in Fig.~\ref{Main Flow}). 
Slices are quality-predicted using the pre-trained NR-Net and then selected based on the following conditions:
\begin{inparaenum}
	\item Predicted labels that are identical to those predicted in the previous iteration;
	\item Predicted labels with high-confidence, i.e., maximal probabilities beyond a threshold. 
\end{inparaenum}
The labels of the selected slices are replaced with the predicted labels.
Slices that do not meet these criteria are pruned from the training dataset. 
NR-Net is then retrained for the next iteration until accuracy improvement is minimal.

\subsection{Volume Self-Training}
Random forest, effective even with small training datasets \cite{Breiman2001Random}, is employed to predict the volumetric quality based on the slice quality ratings.
Both labeled and unlabeled volumes are utilized to train the random forest. The initial quality ratings of unlabeled volumes are determined based on the following rules:
\begin{inparaenum}
	\item ``Pass'' if more than 80 percent of the slices in the volume are labeled as ``pass'';
	\item ``Fail'' if more slices are labeled as ``fail'' than ``pass'' or ``questionable''; 
	\item ``Questionable'' if otherwise. 
\end{inparaenum}

Similar to slice self-training, volume self-training involves iterative volume relabeling/pruning,  random forest retraining (blue box in Fig.~\ref{Main Flow}). 
The input to the random forest is the slice quality ratings predicted using the NR-Net. 
To reduce the influence of label noise, volumes satisfying the following criteria are retained:
\begin{inparaenum}
	\item Predicted labels identical to those predicted in the previous iteration;
 	\item Predicted labels with high-confidence, i.e., maximal probabilities beyond a threshold. 
\end{inparaenum}
The labels of the selected volumes are replaced by the predicted labels.
Volumes that do not meet these criteria are pruned from the training dataset. 
The random forest is then retrained for the next iteration until accuracy improvement is minimal.

\section{Experimental Results}\label{sec:Experiments}
\subsection{Training}\label{sec:preprocess}

\begin{figure}[!t]
	\centering
	\includegraphics[width=8.5cm]{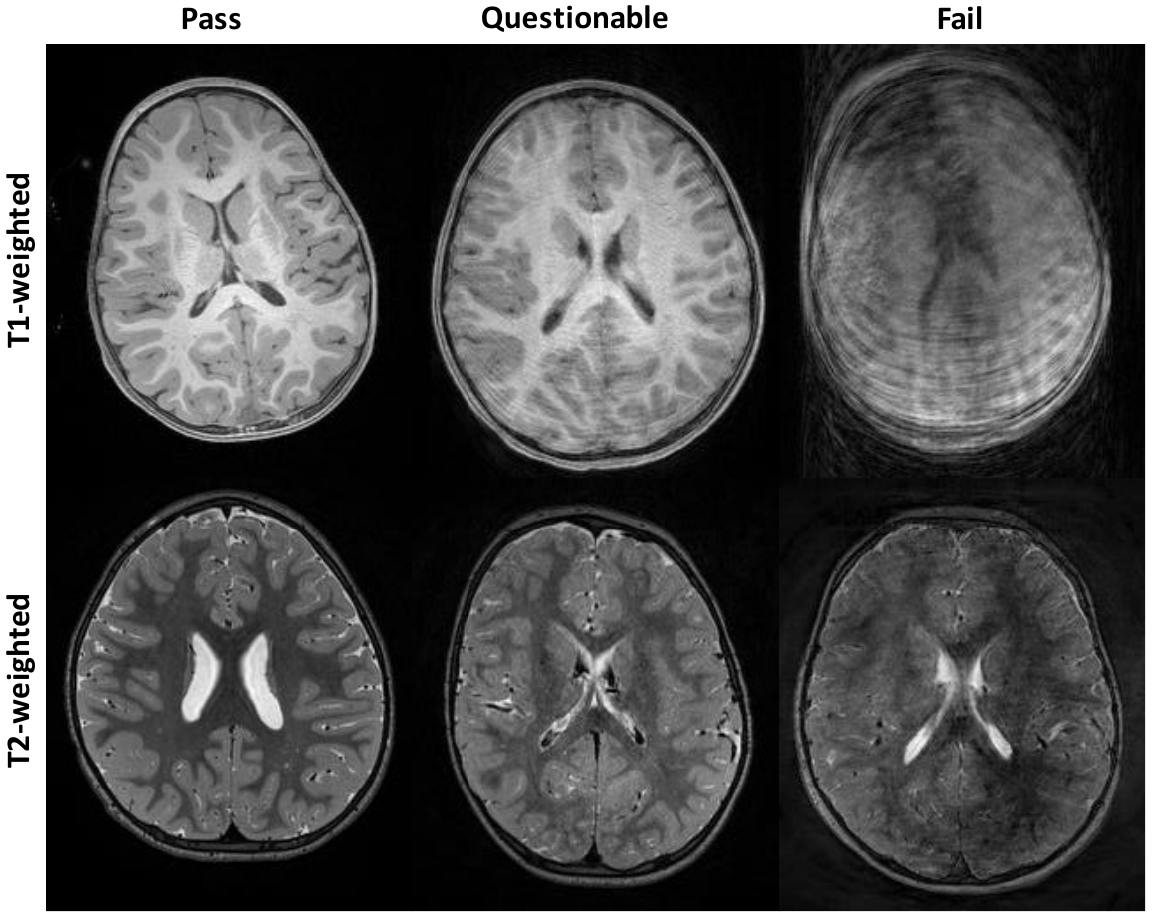}\\
	\caption{Examples of T1- and T2-weighted slices labeled as ``pass'' (no/minor artifacts), ``questionable'' (moderate artifacts), and ``fail'' (heavy artifacts).}
	\label{fig:SampleImages}
\end{figure}

We evaluated our automatic IQA framework on T1- and T2-weighted MR images of pediatric subjects from birth to six years of age \cite{Howell2019The}. The images were separated into three datasets: 
\begin{inparaenum}
	\item Training dataset with noisy labels (annotated volume-wise by an expert);
	\item Testing dataset with reliable labels (annotated volume-wise by multiple experts); 
	and 
  \item Unlabeled dataset.
\end{inparaenum}
See Table~\ref{tab:dataset information} for a summary. Note that, as in practical scenarios, the unlabeled dataset is much larger. 
Fig.~\ref{fig:SampleImages} shows examples of slices labeled as ``pass'' (no/minor artifacts), ``questionable'' (moderate artifacts), and ``fail'' (heavy artifacts).

In total, 3600, 2400, and 26040 axial slices were extracted, respectively, from the 60 T1-weighted training volumes, 40 T1-weighted testing volumes, and 434 T1-weighted unlabeled volumes. 
On the other hand, 3600, 2400, and 22800 axial slices were extracted, respectively, from the 60 T2-weighted training volumes, 40 T2-weighted testing volume, and 380 T2-weighted unlabeled volumes. 
Each slice was uniformly padded to 256$\times$256, min-max intensity normalized, and labeled according to the volume it belongs to.
For both slice and volume assessments, the T1-/T2-weighted slice/volume training sets were divided into training and validation subsets with a ratio of 9:1.

\setlength{\abovecaptionskip}{0pt}
\setlength{\belowcaptionskip}{-20pt}

\begin{table}[!t]
	\caption{Datasets for Training and Testing}
	\centering
	\renewcommand{\arraystretch}{1.4}
		\begin{tabular}{|c|c|c|c|c|c|c|c|}
			\hline
			\multirow{3}{*}{\begin{tabular}[c]{@{}c@{}}Data\\ Type\end{tabular}} & \multicolumn{6}{c|}{Labeled Dataset} & \multirow{3}{*}{\begin{tabular}[c]{@{}c@{}}Unlabeled\\ Dataset\end{tabular}} \\ \cline{2-7}
			& \multicolumn{3}{c|}{Training Dataset} & \multicolumn{3}{c|}{Testing Dataset} &  \\ \cline{2-7}
			& Pass & Ques & Fail & Pass & Ques & Fail &  \\ \hline
			T1w & 20 & 20 & 20 & 25 & 9 & 6 & 434 \\ \hline
			T2w & 20 & 20 & 20 & 21 & 6 & 13 & 380 \\ \hline
	\end{tabular}
	\label{tab:dataset information}
\end{table}

To implement NR-Net, we employed Keras with Tensorflow backend. 
To avoid overfitting, the data were augmented via rotation and horizontal flipping. In addition, $L_2$ regularization was used for the Conv and DSConv layers. 
To deal with data imbalance, a multi-class balanced focal loss \cite{Lin2017Focal} with $L_2$ regularization was used:
\begin{equation}
\mathcal{L}(p_t)= -\alpha_t(1-p_t)^\kappa\log(p_t) + \frac{\lambda}{2n_w}\sum_w \|w\|_2^2,
\label{eq:focalloss}
\end{equation}
where $p_t$, $t=1,2,3$, are the predicted probabilities for ``pass'', ``questionable'', and ``fail''. $\kappa\geq0$ is a focusing parameter, $w$'s are the weight matrices of NR-Net, $\lambda=0.01$ is a tuning parameter for $L_2$ regularization, and $n_w$ is the number of weight matrices. 
Here, the class weights $\alpha_t=\max(N_1,N_2,N_3)/N_{t}$ are used for balancing the contributions of imbalanced datasets. $N_t$ is the number of slices in association with the $t$-th class.
The RMSprop optimizer was employed to learn the network weights, with the initial learning rate set to $1\times10^{-5}$ and the decay rate set to $5\times10^{-8}$. 
Slice self-training was repeated twice.

Random forest for volume prediction was implemented using Scikit-Learn. 
The random forest consisted of 50 trees with entropy as a measure of quality. Balanced class weights were used to counter volume data imbalance. 
Volume self-training was repeated twice.

\subsection{Determination of Thresholds}\label{subsec:ThresholdDetermination}
Utilizing semi-supervised learning (Section~\ref{sec:preprocess}), we obtained 29640 and 26400 T1-weighted and T2-weighted slices labeled with ``pass'', ``questionable'' or ``fail''. These slices were then used for slice and volume self-training, considering 4 probability thresholds $p_{\text{slice}}$, $p_{\text{volume}} \in\{0.6, 0.7, 0.8, 0.9\}$.
Table~\ref{tab:threshold} shows that the prediction accuracy is stable when the threshold is 0.7 and greater.
Thus, in subsequent experiments, we set both $p_{\text{sliice}}$ and $p_{\text{volume}}$ to 0.8. 

\begin{table}[!t]
	\caption{Comparison of different thresholds}
	\centering
	\renewcommand{\arraystretch}{1.4}
	\subtable[Threshold Comparison for T1-Weighted Images]{
		\centering
			\begin{tabular}{|c|c|c|c|c|}
				\hline
				\multirow{2}{*}{$p_{\text{slice}}$} & \multicolumn{4}{c|}{$p_{\text{volume}}$} \\ \cline{2-5} 
				& 0.6     & 0.7     & 0.8     & 0.9     \\ \hline
				0.6                                                                        & 0.8750  & 0.9000  & 0.9000  & 0.8750   \\ \hline
				0.7                                                                        & 0.9750   & 1.0000  & 1.0000  & 1.0000  \\ \hline
				0.8                                                                        & 0.9750   & 1.0000  & 1.0000  & 1.0000  \\ \hline
				0.9                                                                        & 0.9750   & 1.0000  & 1.0000  & 1.0000  \\ \hline
		\end{tabular}
		\label{tab:T1 th}}
	\subtable[Threshold Comparison for T2-Weighted Images]{
		\centering
			\begin{tabular}{|c|c|c|c|c|}
				\hline
				\multirow{2}{*}{$p_{\text{slice}}$} & \multicolumn{4}{c|}{$p_{\text{volume}}$} \\ \cline{2-5} 
				& 0.6     & 0.7     & 0.8     & 0.9     \\ \hline
				0.6                                                                        & 0.8250  & 0.8250  & 0.8750  & 0.8750   \\ \hline
				0.7                                                                        & 0.9250   & 1.0000  & 1.0000  & 1.0000  \\ \hline
				0.8                                                                        & 0.9250   & 1.0000  & 1.0000  & 1.0000  \\ \hline
				0.9                                                                        & 0.9250   & 1.0000  & 1.0000  & 1.0000  \\ \hline
		\end{tabular}
		\label{tab:T2 th}}
	\label{tab:threshold}
\end{table}

\subsection{Ablation Study}
To verify the effectiveness of DSRes and NRes blocks, we compared modified versions of NR-Net:
\begin{itemize}
  \item Convolution residual (CRes) network, which substitutes the DSRes blocks and NRes block with CRes blocks with channel numbers 128, 256 and 512;
  \item CRes+NRes network, which substitutes DSRes blocks with CRes blocks with channel numbers 128 and 256;
  \item DSRes network, which substitutes the NRes block with DSRes block with channel number 512. 
\end{itemize} 
Note that NR-Net is a DSRes+NRes network. 
These networks were trained in a manner similar to the NR-Net.

\begin{table}[!t]
	\caption{Number of parameters (NoP), maximal dimension (MD), and time cost (TC) on GPU and CPU.}
	\centering
	\renewcommand{\arraystretch}{1.5}
	\resizebox{0.49\textwidth}{!}{
		\begin{tabular}{|c|c|c|c|c|c|c|}
			\hline
			\multirow{2}{*}{Network} & \multirow{2}{*}{NoP} & \multirow{2}{*}{MD} & \multicolumn{2}{c|}{TC (Slice)} & \multicolumn{2}{c|}{TC (Volume)} \\ \cline{4-7} 
			&                         &                        & GPU             & CPU             & GPU            & CPU              \\ \hline
			CRes              & 4.86M                   & 512                    & 10.71ms         & 0.228s         & 355ms           & 11.635s          \\ \hline
			CRes+NRes         & 1.31M                   & 256                    & 10.35ms         & 0.197s         & 333ms           & 10.251s          \\ \hline
			DSRes             & 0.75M                   & 512                    & 10.58ms         & 0.175s         & 347ms           & 7.874s           \\ \hline
			DSRes+NRes        & 0.33M                   & 256                    & 10.01ms         & 0.159s         & 312ms           & 6.723s           \\ \hline
	\end{tabular}}
	\label{tab:time cost}
\end{table}

\begin{table*}[!t]
	\caption{Confusion matrices of CRes, CRes+NRes, DSRes and DSRes+NRes for T1-weighted images}
	\centering
	\renewcommand{\arraystretch}{1.3}
	\subtable[CRes]{
		\centering
		\resizebox{0.49\textwidth}{!}{
			\begin{tabular}{|c|c|c|c|c|c|c|c|}
				\hline
				\multicolumn{2}{|c|}{\multirow{3}{*}{\begin{tabular}[c]{@{}c@{}}Image\\ Quality\end{tabular}}} & \multicolumn{6}{c|}{Predicted} \\ \cline{3-8}
				\multicolumn{2}{|c|}{} & \multicolumn{2}{c|}{Pass} & \multicolumn{2}{c|}{Ques} & \multicolumn{2}{c|}{Fail}\\ \cline{3-8} 
				\multicolumn{2}{|c|}{} & Slice & Volume & Slice & Volume & Slice & Volume \\ \hline
				\multirow{3}{*}{\rotatebox{90}{Actual}} & Pass & 1421 & 23 & ~79~ & 2 & 0 & 0\\ \cline{2-8} 
				& Ques & - & 0 & - & 9 & - & 0\\ \cline{2-8}
				& Fail & 0 & 0 & 3 & 0 & 357 & 6\\ \hline
		\end{tabular}}\label{tab:T1-cres-conmtx}}
	\subtable[CRes+NRes]{
		\centering
		\resizebox{0.49\textwidth}{!}{
			\begin{tabular}{|c|c|c|c|c|c|c|c|}
				\hline
				\multicolumn{2}{|c|}{\multirow{3}{*}{\begin{tabular}[c]{@{}c@{}}Image\\ Quality\end{tabular}}} & \multicolumn{6}{c|}{Predicted} \\ \cline{3-8}
				\multicolumn{2}{|c|}{} & \multicolumn{2}{c|}{Pass} & \multicolumn{2}{c|}{Ques} & \multicolumn{2}{c|}{Fail}\\ \cline{3-8} 
				\multicolumn{2}{|c|}{} & Slice & Volume & Slice & Volume & Slice & Volume \\ \hline
				\multirow{3}{*}{\rotatebox{90}{Actual}} & Pass & 1470 & 24 & 30 & 1 & 0 & 0\\ \cline{2-8} 
				& Ques & - & 0 & - & 9 & - & 0\\ \cline{2-8} 
				& Fail & 0 & 0 & 6 & 0 & 354 & 6\\ \hline
		\end{tabular}}\label{tab:T1-cres+nres-conmtx}}
	\subtable[DSRes]{
		\centering
		\resizebox{0.49\textwidth}{!}{
			\begin{tabular}{|c|c|c|c|c|c|c|c|}
				\hline
				\multicolumn{2}{|c|}{\multirow{3}{*}{\begin{tabular}[c]{@{}c@{}}Image\\ Quality\end{tabular}}} & \multicolumn{6}{c|}{Predicted} \\ \cline{3-8}
				\multicolumn{2}{|c|}{} & \multicolumn{2}{c|}{Pass} & \multicolumn{2}{c|}{Ques} & \multicolumn{2}{c|}{Fail}\\ \cline{3-8} 
				\multicolumn{2}{|c|}{} & Slice & Volume & Slice & Volume & Slice & Volume \\ \hline
				\multirow{3}{*}{\rotatebox{90}{Actual}} & Pass & 1437 & 24 & 62 & 1 & 1 & 0\\ \cline{2-8} 
				& Ques & - & 0 & - & 9 & - & 0\\ \cline{2-8} 
				& Fail & 0 & 0 & 5 & 0 & 355 & 6\\ \hline
		\end{tabular}}\label{tab:T1-dsres-conmtx}}
	\subtable[DSRes+NRes]{
		\centering
		\resizebox{0.49\textwidth}{!}{
			\begin{tabular}{|c|c|c|c|c|c|c|c|}
				\hline
				\multicolumn{2}{|c|}{\multirow{3}{*}{\begin{tabular}[c]{@{}c@{}}Image\\ Quality\end{tabular}}} & \multicolumn{6}{c|}{Predicted} \\ \cline{3-8}
				\multicolumn{2}{|c|}{} & \multicolumn{2}{c|}{Pass} & \multicolumn{2}{c|}{Ques} & \multicolumn{2}{c|}{Fail}\\ \cline{3-8} 
				\multicolumn{2}{|c|}{} & Slice & Volume & Slice & Volume & Slice & Volume \\ \hline
				\multirow{3}{*}{\rotatebox{90}{Actual}} & Pass & 1500 & 25 & 0 & 0 & 0 & 0\\ \cline{2-8} 
				& Ques & - & 0 & - & 9 & - & 0\\ \cline{2-8} 
				& Fail & 0 & 0 & 56 & 0 & 304 & 6\\ \hline
		\end{tabular}}\label{tab:T1-dsres+nres-conmtx}}
	\label{tab:T1-conmtx}
\end{table*}

\begin{table*}[!t]
	\caption{Sensitivity and specificity of CRes, CRes+NRes, DSRes and DSRes+NRes for T1-weighted images}
	\centering
	\renewcommand{\arraystretch}{1.4}
	\subtable[CRes]{
		\centering
		\resizebox{0.4\textwidth}{!}{
			\begin{tabular}{|c|c|c|c|c|}
				\hline
				\multirow{2}{*}{Image Quality} & \multicolumn{2}{c|}{Sensitivity} & \multicolumn{2}{c|}{Specificity} \\ \cline{2-5}
				& Slice & Volume & Slice & Volume\\ \hline
				Pass & 0.9473 & 0.9200 & 1.0000 & 1.0000 \\ \cline{1-5} 
				Ques & - & 1.0000 & - & 0.9355 \\ \cline{1-5} 
				Fail & 0.9917 & 1.0000 & 1.0000 & 1.0000 \\ \hline
		\end{tabular}}\label{tab:T1-cres-senspe}}
	\subtable[CRes+NRes]{
		\centering
		\resizebox{0.4\textwidth}{!}{
			\begin{tabular}{|c|c|c|c|c|}
				\hline
				\multirow{2}{*}{Image Quality} & \multicolumn{2}{c|}{Sensitivity} & \multicolumn{2}{c|}{Specificity} \\ \cline{2-5}
				& Slice & Volume & Slice & Volume\\ \hline
				Pass & 0.9800 & 0.9600 & 1.0000 & 1.0000 \\ \cline{1-5} 
				Ques & - & 1.0000 & - & 0.9667 \\ \cline{1-5} 
				Fail & 0.9833 & 1.0000 & 1.0000 & 1.0000 \\ \hline
		\end{tabular}}\label{tab:T1-cres+nres-senspe}}
	\subtable[DSRes]{
		\centering
		\resizebox{0.4\textwidth}{!}{
			\begin{tabular}{|c|c|c|c|c|}
				\hline
				\multirow{2}{*}{Image Quality} & \multicolumn{2}{c|}{Sensitivity} & \multicolumn{2}{c|}{Specificity} \\ \cline{2-5}
				& Slice & Volume & Slice & Volume\\ \hline
				Pass & 0.9580 & 0.9600 & 1.0000 & 1.0000 \\ \cline{1-5} 
				Ques & - & 1.0000 & - & 0.9677 \\ \cline{1-5} 
				Fail & 0.9861 & 1.0000 & 0.9993 & 1.0000 \\ \hline
		\end{tabular}}\label{tab:T1-dsres-senspe}}
	\subtable[DSRes+NRes]{
		\centering
		\resizebox{0.4\textwidth}{!}{
			\begin{tabular}{|c|c|c|c|c|}
				\hline
				\multirow{2}{*}{Image Quality} & \multicolumn{2}{c|}{Sensitivity} & \multicolumn{2}{c|}{Specificity} \\ \cline{2-5}
				& Slice & Volume & Slice & Volume\\ \hline
				Pass & 1.0000 & 1.0000 & 1.0000 & 1.0000 \\ \cline{1-5} 
				Ques & - & 1.0000 & - & 1.0000 \\ \cline{1-5} 
				Fail & 0.8444 & 1.0000 & 1.0000 & 1.0000 \\ \hline
		\end{tabular}}\label{tab:T1-dsres+nres-senspe}}
	\label{tab:T1-senspe}
\end{table*}

\begin{figure*}[!t]
	\centering
	\includegraphics[width=18cm]{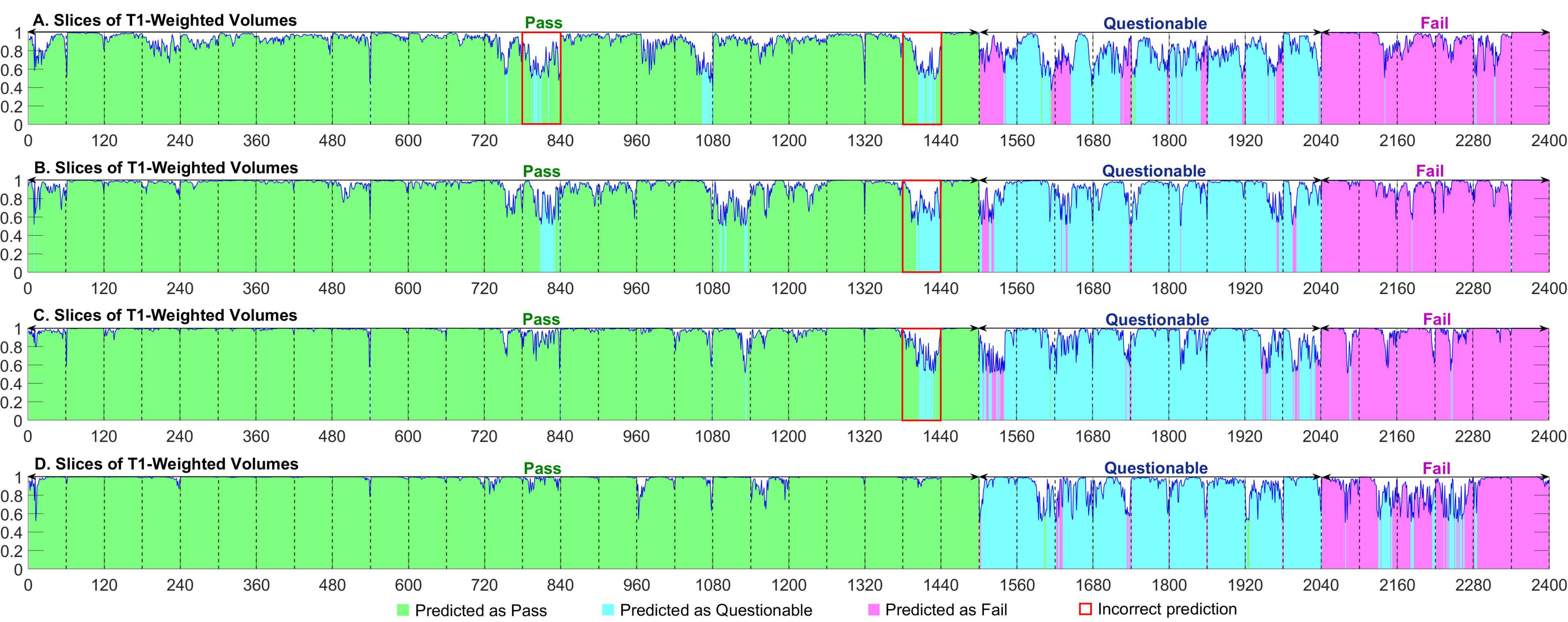}\\
	\caption{Quality assessment of T1-weighted images. (A) CRes, (B) CRes+NRes, (C) DSRes, and (D) DSRes+NRes. The slices of each volume are marked by dashed vertical lines.}
	\label{fig:T1Comparison}
\end{figure*}

\begin{table*}[!t]
	\caption{Confusion matrices of CRes, CRes+NRes, DSRes and DSRes+NRes for T2-weighted images}
	\centering
	\renewcommand{\arraystretch}{1.3}
	\subtable[CRes]{
		\centering
		\resizebox{0.49\textwidth}{!}{
			\begin{tabular}{|c|c|c|c|c|c|c|c|}
				\hline
				\multicolumn{2}{|c|}{\multirow{3}{*}{\begin{tabular}[c]{@{}c@{}}Image\\ Quality\end{tabular}}} & \multicolumn{6}{c|}{Predicted} \\ \cline{3-8}
				\multicolumn{2}{|c|}{} & \multicolumn{2}{c|}{Pass} & \multicolumn{2}{c|}{Ques} & \multicolumn{2}{c|}{Fail}\\ \cline{3-8} 
				\multicolumn{2}{|c|}{} & Slice & Volume & Slice & Volume & Slice & Volume \\ \hline
				\multirow{3}{*}{\rotatebox{90}{Actual}} & Pass & 1243 & 21 & 12 & 0 & 5 & 0\\ \cline{2-8} 
				& Ques & - & 1 & - & 4 & - & 1\\ \cline{2-8}
				& Fail & 0 & 0 & 0 & 0 & 780 & 13\\ \hline
		\end{tabular}}\label{tab:T2-cres-conmtx}}
	\subtable[CRes+NRes]{
		\centering
		\resizebox{0.49\textwidth}{!}{
			\begin{tabular}{|c|c|c|c|c|c|c|c|}
				\hline
				\multicolumn{2}{|c|}{\multirow{3}{*}{\begin{tabular}[c]{@{}c@{}}Image\\ Quality\end{tabular}}} & \multicolumn{6}{c|}{Predicted} \\ \cline{3-8}
				\multicolumn{2}{|c|}{} & \multicolumn{2}{c|}{Pass} & \multicolumn{2}{c|}{Ques} & \multicolumn{2}{c|}{Fail}\\ \cline{3-8} 
				\multicolumn{2}{|c|}{} & Slice & Volume & Slice & Volume & Slice & Volume \\ \hline
				\multirow{3}{*}{\rotatebox{90}{Actual}} & Pass & 1257 & 21 & 3 & 0 & 0 & 0\\ \cline{2-8} 
				& Ques & - & 2 & - & 4 & - & 0\\ \cline{2-8} 
				& Fail & 0 & 0 & 6 & 0 & 774 & 13\\ \hline
		\end{tabular}}\label{tab:T2-cres+nres-conmtx}}
	\subtable[DSRes]{
		\centering
		\resizebox{0.49\textwidth}{!}{
			\begin{tabular}{|c|c|c|c|c|c|c|c|}
				\hline
				\multicolumn{2}{|c|}{\multirow{3}{*}{\begin{tabular}[c]{@{}c@{}}Image\\ Quality\end{tabular}}} & \multicolumn{6}{c|}{Predicted} \\ \cline{3-8}
				\multicolumn{2}{|c|}{} & \multicolumn{2}{c|}{Pass} & \multicolumn{2}{c|}{Ques} & \multicolumn{2}{c|}{Fail}\\ \cline{3-8} 
				\multicolumn{2}{|c|}{} & Slice & Volume & Slice & Volume & Slice & Volume \\ \hline
				\multirow{3}{*}{\rotatebox{90}{Actual}} & Pass & 1242 & 21 & 15 & 0 & 3 & 0\\ \cline{2-8} 
				& Ques & - & 1 & - & 3 & - & 2\\ \cline{2-8} 
				& Fail & 0 & 0 & 6 & 0 & 774 & 13\\ \hline
		\end{tabular}}\label{tab:T2-dsres-conmtx}}
	\subtable[DSRes+NRes]{
		\centering
		\resizebox{0.49\textwidth}{!}{
			\begin{tabular}{|c|c|c|c|c|c|c|c|}
				\hline
				\multicolumn{2}{|c|}{\multirow{3}{*}{\begin{tabular}[c]{@{}c@{}}Image\\ Quality\end{tabular}}} & \multicolumn{6}{c|}{Predicted} \\ \cline{3-8}
				\multicolumn{2}{|c|}{} & \multicolumn{2}{c|}{Pass} & \multicolumn{2}{c|}{Ques} & \multicolumn{2}{c|}{Fail}\\ \cline{3-8} 
				\multicolumn{2}{|c|}{} & Slice & Volume & Slice & Volume & Slice & Volume \\ \hline
				\multirow{3}{*}{\rotatebox{90}{Actual}} & Pass & 1248 & 21 & 12 & 0 & 0 & 0\\ \cline{2-8} 
				& Ques & - & 0 & - & 6 & - & 0\\ \cline{2-8} 
				& Fail & 0 & 0 & 0 & 0 & 780 & 13\\ \hline
		\end{tabular}}\label{tab:T2-dsres+nres-conmtx}}
	\label{tab:T2-conmtx}
\end{table*}

\begin{table*}[!t]
	\caption{Sensitiviy and specificity of CRes, CRes+NRes, DSRes and DSRes+NRes for T2-weighted images}
	\centering
	\renewcommand{\arraystretch}{1.4}
	\subtable[CRes]{
		\centering
		\resizebox{0.4\textwidth}{!}{
			\begin{tabular}{|c|c|c|c|c|}
				\hline
				\multirow{2}{*}{Image Quality} & \multicolumn{2}{c|}{Sensitivity} & \multicolumn{2}{c|}{Specificity} \\ \cline{2-5}
				& Slice & Volume & Slice & Volume\\ \hline
				Pass & 0.9865 & 1.0000 & 1.0000 & 0.9474 \\ \cline{1-5} 
				Ques & - & 0.6667 & - & 1.0000 \\ \cline{1-5} 
				Fail & 1.0000 & 1.0000 & 0.9960 & 0.9630 \\ \hline
		\end{tabular}}\label{tab:T2-cres-senspe}}
	\subtable[CRes+NRes]{
		\centering
		\resizebox{0.4\textwidth}{!}{
			\begin{tabular}{|c|c|c|c|c|}
				\hline
				\multirow{2}{*}{Image Quality} & \multicolumn{2}{c|}{Sensitivity} & \multicolumn{2}{c|}{Specificity} \\ \cline{2-5}
				& Slice & Volume & Slice & Volume\\ \hline
				Pass & 0.9976 & 1.0000 & 1.0000 & 0.8947 \\ \cline{1-5} 
				Ques & - & 0.6667 & - & 1.0000 \\ \cline{1-5} 
				Fail & 0.9923 & 1.0000 & 1.0000 & 1.0000 \\ \hline
		\end{tabular}}\label{tab:T2-cres+nres-senspe}}
	\qquad
	\subtable[DSRes]{
		\centering
		\resizebox{0.4\textwidth}{!}{
			\begin{tabular}{|c|c|c|c|c|}
				\hline
				\multirow{2}{*}{Image Quality} & \multicolumn{2}{c|}{Sensitivity} & \multicolumn{2}{c|}{Specificity} \\ \cline{2-5}
				& Slice & Volume & Slice & Volume\\ \hline
				Pass & 0.9857 & 1.0000 & 1.0000 & 0.9474 \\ \cline{1-5} 
				Ques & - & 0.5000 & - & 1.0000 \\ \cline{1-5} 
				Fail & 0.9923 & 1.0000 & 0.9976 & 0.9259 \\ \hline
		\end{tabular}}\label{tab:T2-dsres-senspe}}
	\subtable[DSRes+NRes]{
		\centering
		\resizebox{0.4\textwidth}{!}{
			\begin{tabular}{|c|c|c|c|c|}
				\hline
				\multirow{2}{*}{Image Quality} & \multicolumn{2}{c|}{Sensitivity} & \multicolumn{2}{c|}{Specificity} \\ \cline{2-5}
				& Slice & Volume & Slice & Volume\\ \hline
				Pass & 0.9905 & 1.0000 & 1.0000 & 1.0000 \\ \cline{1-5} 
				Ques & - & 1.0000 & - & 1.0000 \\ \cline{1-5} 
				Fail & 1.0000 & 1.0000 & 1.0000 & 1.0000 \\ \hline
		\end{tabular}}\label{tab:T2-dsres+nres-senspe}}
	\label{tab:T2-senspe}
\end{table*}

\begin{figure*}[!t]
	\centering
	\includegraphics[width=18cm]{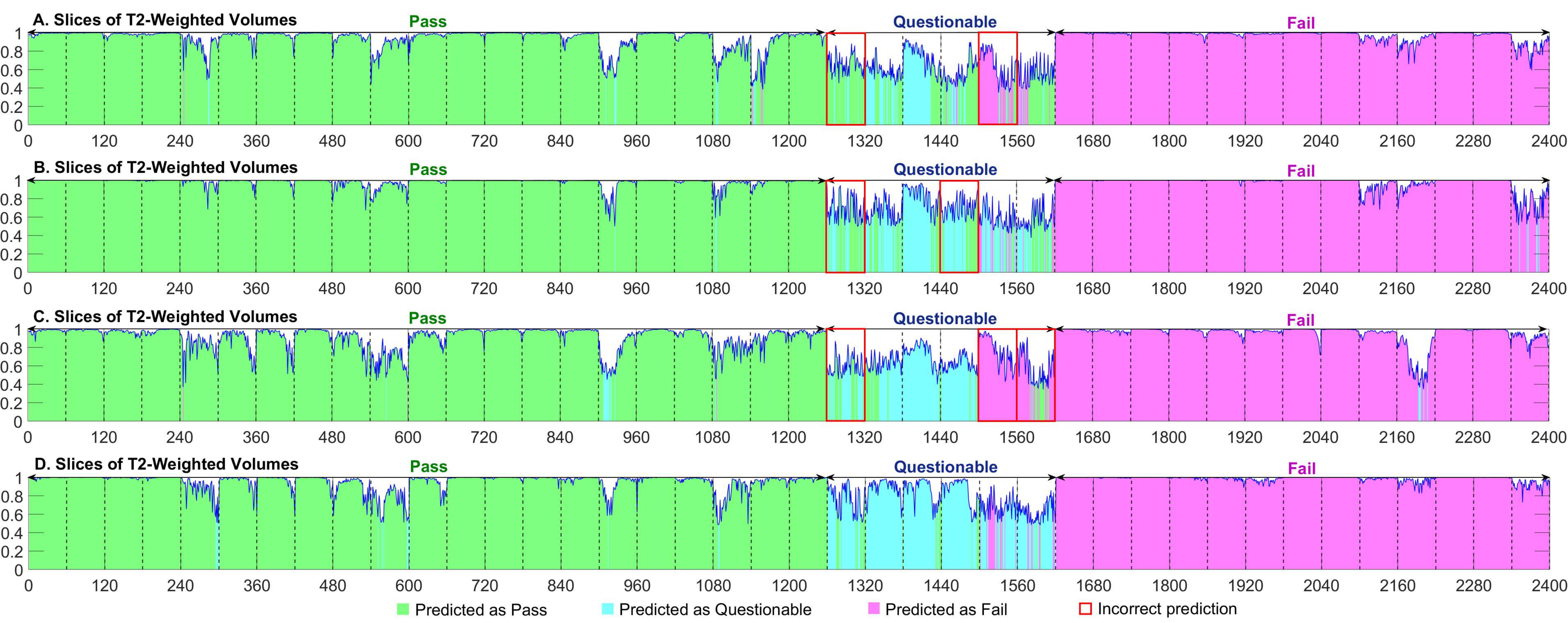}\\
	\caption{Quality assessment of T2-weighted images. (A) CRes, (B) CRes+NRes, (C) DSRes, and (D) DSRes+NRes. The slices of each volume are marked by dashed vertical lines.  
	}
	\label{fig:T2Comparison}
\end{figure*}

\begin{figure*}[!t]
	\centering
	\subfigure[``Questionable'' slices]{
		\includegraphics[width=0.30\textwidth]{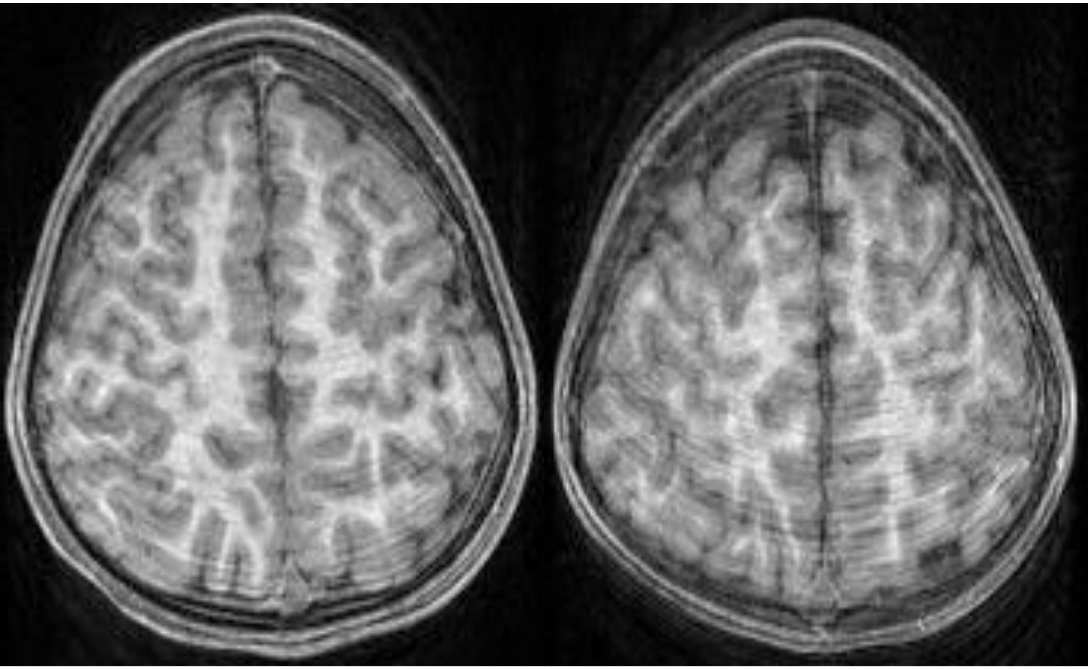}\label{fig:T1ques}}
	\subfigure[``Fail'' slices]{
		\includegraphics[width=0.30\textwidth]{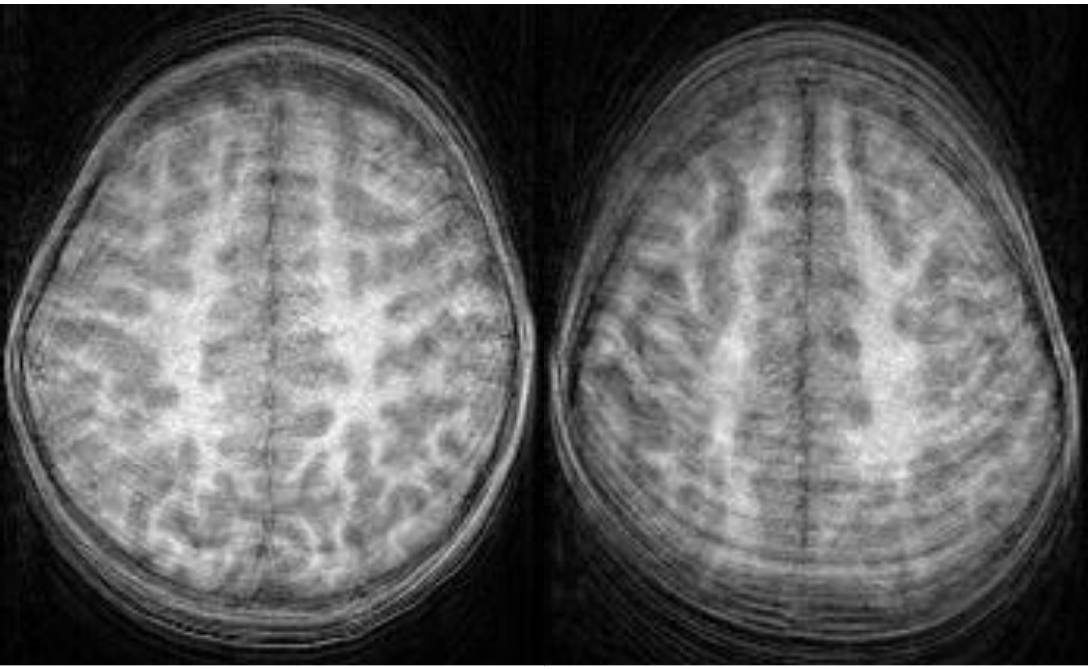}\label{fig:T1fail}}
	\subfigure[Mispredicted slices]{
		\includegraphics[width=0.30\textwidth]{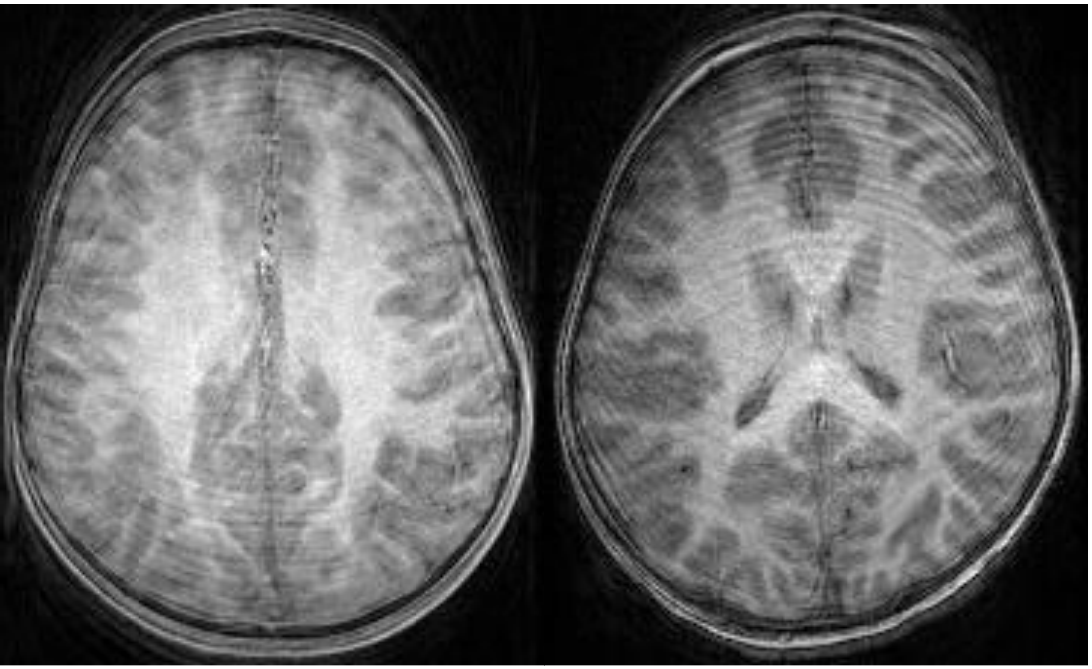}\label{fig:T1mis}}
	\caption{Examples of ``questionable'', ``fail'', and mispredicted T1-weighted slices.}
	\label{fig:T1DCA}
\end{figure*}
\begin{figure*}[!t]
	\centering
	\subfigure[``Pass'' slices]{
		\includegraphics[width=0.30\textwidth]{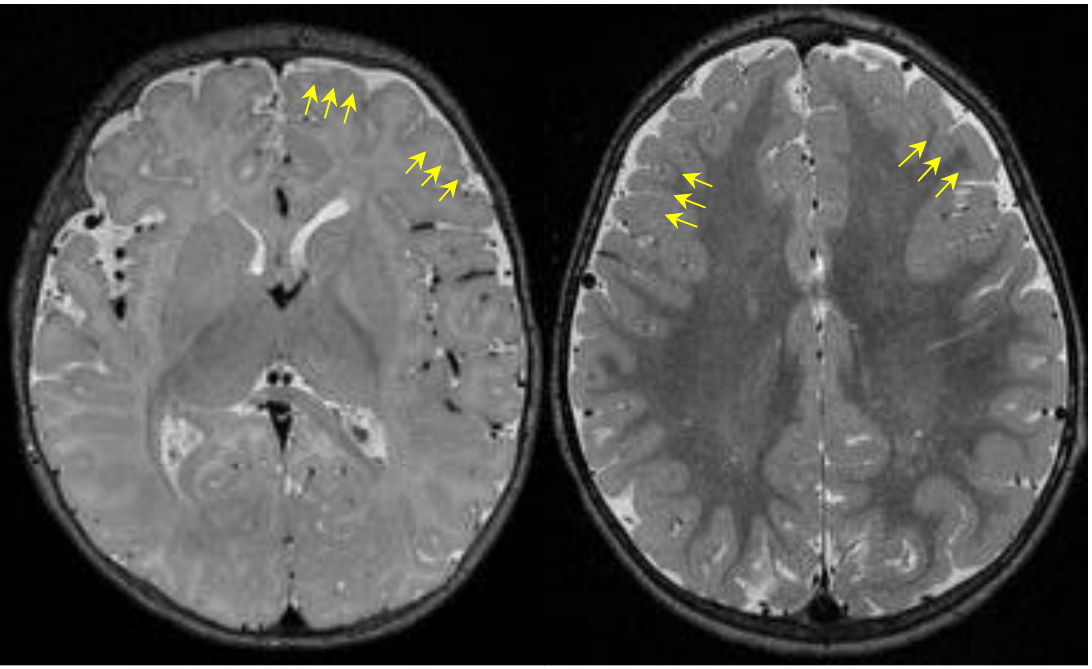}\label{fig:T2pass}}
	\subfigure[``Questionable'' slices]{
		\includegraphics[width=0.30\textwidth]{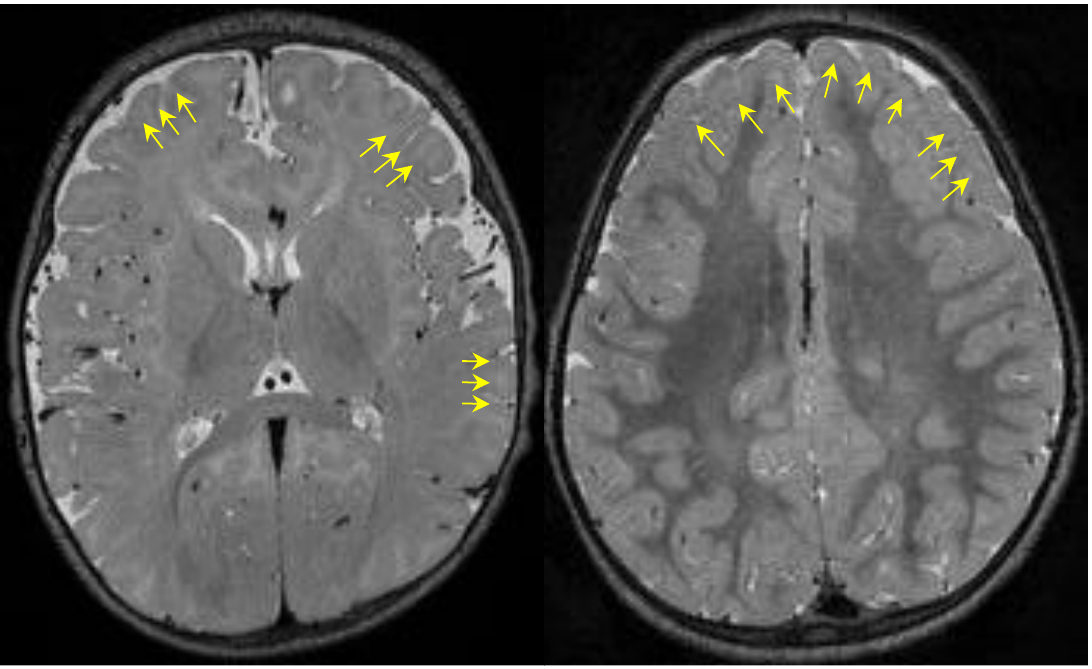}\label{fig:T2ques}}
	\centering
	\subfigure[Mispredicted slices]{
		\includegraphics[width=0.30\textwidth]{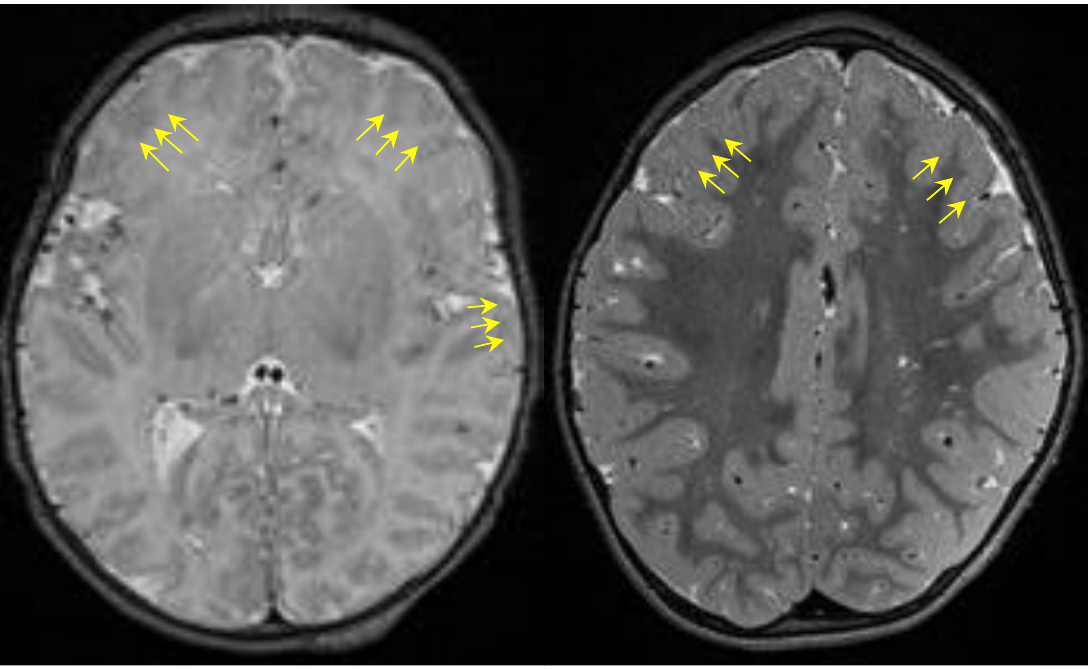}\label{fig:T2mis}}
	\caption{Examples of ``pass'', ``questionable'', and mispredicted T2-weighted slices. The arrows mark some artifacts.}
	\label{fig:T2DCA}
\end{figure*}
\subsubsection{Computational Efficiency}
We compared the computational efficiency of NR-Net (DSRes+NRes network) with the other networks using several metrics: Number of parameters (NoP), the maximal dimension (MD), GPU and CPU time costs (TCs) of slice and volume. An NVIDIA GeForce GTX 1080Ti (GPU) and an Intel i7-8700K (CPU) were used for evaluation.

As shown in Table~\ref{tab:time cost}, the NoP and TCs of NR-Net (DSRes+NRes network) are much smaller than the other networks (e.g., 14$\times$ times smaller than CRes) and the CPU TC of each slice and volume is $\sim$1.5 times lower. 
Compared with CRes and CRes+NRes, the NRes block reduces the maximal dimension from 512 to 256 and reduces the NoP by over three times. 
The GPU and CPU TCs of slice and volume are reduced by over 3\%, and particularly both the CPU slice and volume TC are reduced by over 11.8\%.
Compared with the NoP and TCs of DSRes, the NoP of DSRes+NRes is reduced by over 2 times. The GPU and CPU TCs of slice and volume are all reduced by over 5\%, particularly the CPU volume TC, which is reduced by over 15\%.
CRes and DSRes have the same maximal dimension. The NoP of DSRes is over 6 times smaller than CRes. 
Comparing the TCs of CRes and DSRes, the reduction of TCs on GPU is only $\sim$2\%, but on CPU it is quite significant, i.e., $\sim$23\% and $\sim$33\% reduction of slice and volume TCs, respectively. 
Similarly, comparing CRes+NRes and DSRes+NRes, the NoP reduction reaches almost 4 times and the reduction of GPU and CPU TCs reaches over 3\% and 20\%, respectively.
The analysis above shows that DSRes block and NRes block improve the computational efficiency of the overall network, making it suitable for real-time IQA.

\subsubsection{Network Efficiency}
Tables~\ref{tab:T1-conmtx} and \ref{tab:T1-senspe} show the confusion matrices, along with the sensitivity and specificity of the different methods for T1-weighted testing images. The corresponding results for T2-weighted testing images are shown in Tables~\ref{tab:T2-conmtx} and \ref{tab:T2-senspe}. 
The detailed IQA results for the testing T1- and T2-weighted images are shown in Figs.~\ref{fig:T1Comparison} and \ref{fig:T2Comparison}, respectively.

It can be observed from Tables~\ref{tab:T1-conmtx} and \ref{tab:T1-senspe} that the proposed method yields the best volume IQA performance than the other methods in terms of sensitivity and specificity. 
The slice prediction results in Tables~\ref{tab:T1-conmtx} and \ref{tab:T1-senspe} and Fig. \ref{fig:T1Comparison} show that the sensitivity of ``fail'' slices for DSRes+NRes method is lower than the other methods. This however does not affect the sensitivity of volume IQA.
Similar conclusions can be drawn for T2-weighted images from Tables~\ref{tab:T2-conmtx} and \ref{tab:T2-senspe} and Fig.~\ref{fig:T2Comparison}, showing that 
the proposed method consistently yields the best performance for volume IQA.

\subsection{Discordant Case Analysis}
It can be observed from Table~\ref{tab:T1-dsres+nres-conmtx} and Fig.~\ref{fig:T1Comparison}-D that ``fail'' T1-weighted slices can be mistakenly predicted as ``questionable''. 
By retrospectively inspecting all ``questionable'' and ``fail'' T1-weighted training slices, we found that this confusion is caused by the existence of ringing and motion artifacts in most ``questionable'' and ``fail'' slices, as shown in Fig.~\ref{fig:T1DCA}. 
Similarly, it can be observed from Table~\ref{tab:T2-dsres+nres-conmtx} and Fig.~\ref{fig:T2Comparison}-D that ``pass'' T2-weighted slices can sometimes be mispredicted as ``questionable''. Subtle degradation such as contrast reduction and local fuzziness can confuse the IQA network and causes mispredictions as shown in Fig.~\ref{fig:T2DCA}. 

\begin{figure*}[!t]
	\centering
	\subfigure[Slice-wise IQA for T1-weighted images]{\includegraphics[width=0.4\textwidth]{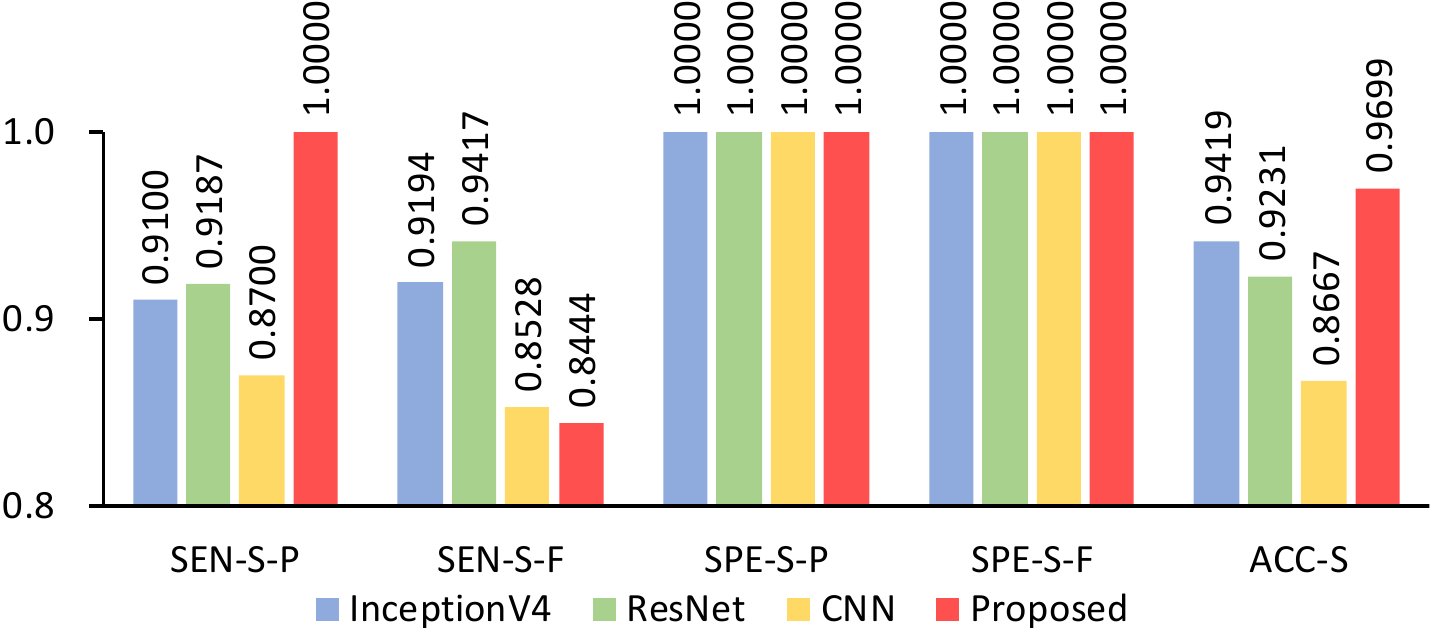}}
	\subfigure[Volume-wise IQA for T1-weighted images]{\includegraphics[width=0.53\textwidth]{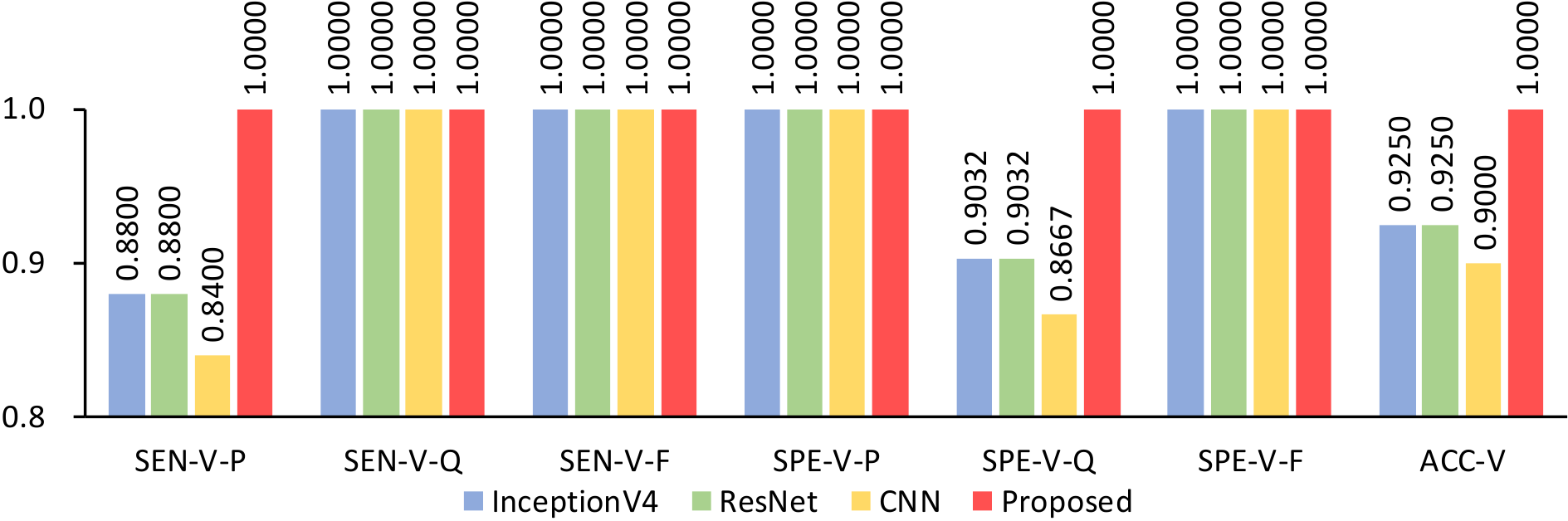}}
	\caption{Slice- and volume-wise IQA for T1-weighted images. Evaluation metrics are denoted in the following form: metric (SEN/SPE/ACC) - stage (S/V) - quality (P/Q/F/-).}
	\label{fig:T1_comp}
\end{figure*}
\begin{figure*}[!t]
	\centering
	\subfigure[Slice-wise IQA for T2-weighted images]{\includegraphics[width=0.39\textwidth]{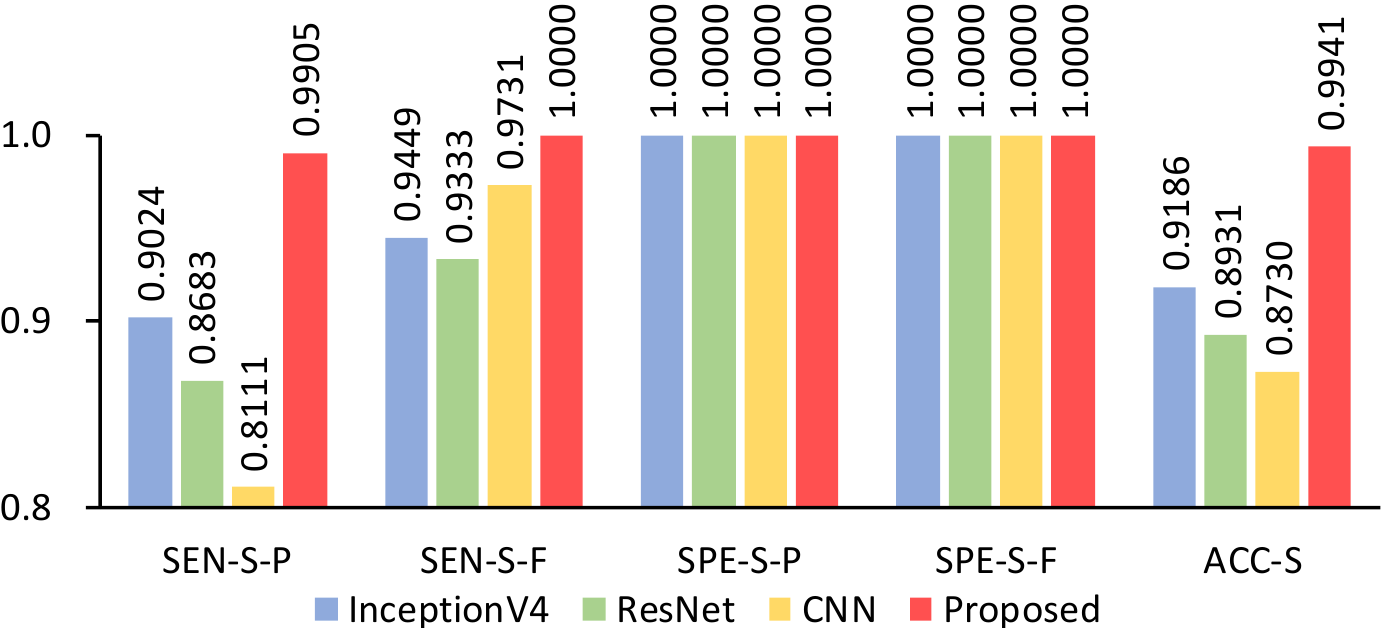}}
	\subfigure[Volume-wise IQA for T2-weighted images]{\includegraphics[width=0.54\textwidth]{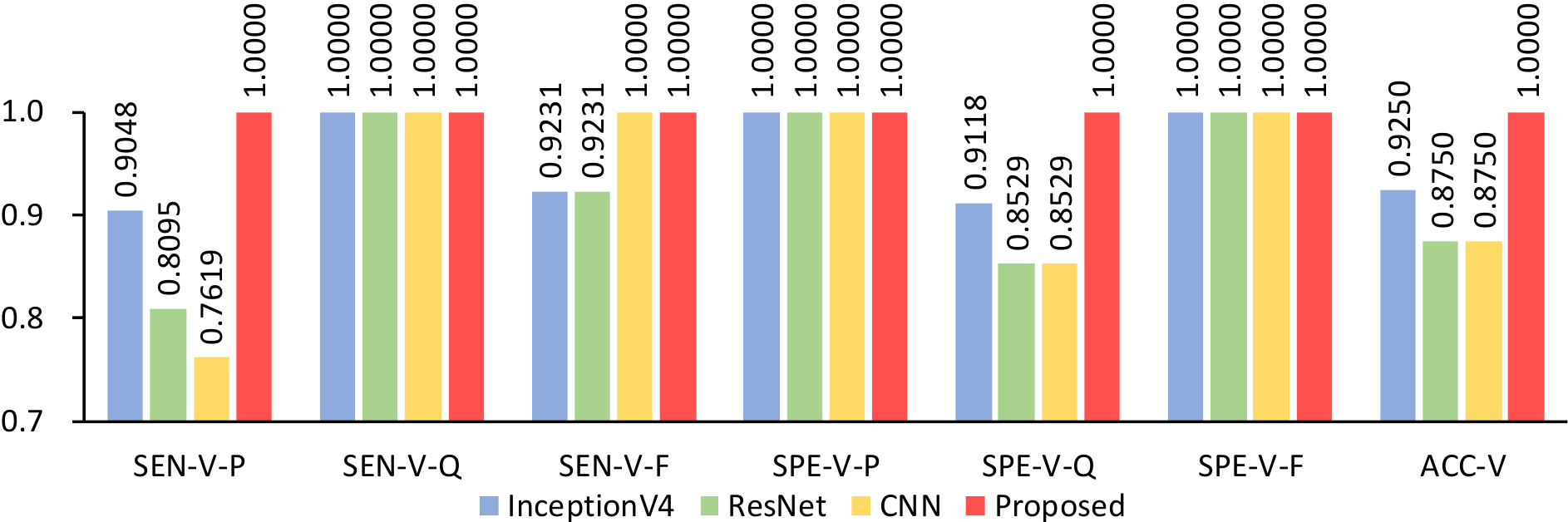}}
	\caption{Slice- and volume-wise IQA for T2-weighted images. Evaluation metrics are denoted in the following form: metric (SEN/SPE/ACC) - stage (S/V) - quality (P/Q/F/-).}
	\label{fig:T2_comp}
\end{figure*}
\subsection{Performance Comparison}
To verify the effectiveness of the proposed IQA method, we compared it with several deep learning IQA methods:
	\begin{itemize}
		\item Pretrained and fine-tuned \cite{Muelly2017Automated}: InceptionV4 \cite{Szegedy2017Inception}, ResNet \cite{He2016Deep}
		\item Trained from scratch: CNN \cite{Esses2017Automated}
	\end{itemize}
	For volume-wise IQA, our random forest is applied to the slice-wise quality ratings given by these methods.

Figs.~\ref{fig:T1_comp} and \ref{fig:T2_comp}, respectively, shows the evaluation results in terms of sensitivity (SEN), specificity (SPE), and accuracy (ACC) for slice- and volume-wise IQA of the T1- and T2-weighted testing data.
From Figs.~\ref{fig:T1_comp} and \ref{fig:T2_comp}, our method significantly outperforms all compared methods for all evalution metrics except sensitivity to ``fail'' T1-weighted slices (i.e., SEN-S-F).

\subsection{Effectiveness of Semi-Supervised Learning}
We evaluated whether semi-supervised learning is able to correctly assess of the large amount of unlabeled data. The unlabeled data were annotated by an expert and the labels were compared with the predictions given by our method.
As can be observed from Table~\ref{tab:noisy}, the automated and predicted volume IQA results are largely consistent. This implies that our method is able to effectively harness the large amount of unlabeled data for increasing sample size and hence improving network training.

\begin{table}[!t]
	\caption{Effectiveness of semi-supervised learning}
	\centering
	\renewcommand{\arraystretch}{1.4}
	\subtable[T1-weighted images]{
		\centering
			\begin{tabular}{|c|c|c|c|c|c|c|c|}
				\hline
				\multicolumn{2}{|c|}{\multirow{2}{*}{Image Quality}} & \multicolumn{3}{c|}{Predicted} & \multirow{2}{*}{Sensitivity} & \multirow{2}{*}{Specificity} \\ \cline{3-5}
				\multicolumn{2}{|c|}{} & Pass & Ques & Fail &  &  \\ \hline
				\multirow{3}{*}{\rotatebox{90}{Label}} & Pass & 319 & 1 & 0 & 0.9969 & 0.9561 \\ \cline{2-7} 
				& Ques & 5 & 79 & 0 & 0.9405 & 0.9971 \\ \cline{2-7} 
				& Fail & 0 & 0 & 30 & 1.0000 & 1.0000 \\ \hline
		\end{tabular}}\label{tab:T1-noisy}
	\subtable[T2-weighted images]{
		\centering
			\begin{tabular}{|c|c|c|c|c|c|c|c|}
				\hline
				\multicolumn{2}{|c|}{\multirow{2}{*}{Image Quality}} & \multicolumn{3}{c|}{Predicted} & \multirow{2}{*}{Sensitivity} & \multirow{2}{*}{Specificity} \\ \cline{3-5}
				\multicolumn{2}{|c|}{} & Pass & Ques & Fail &  &  \\ \hline
				\multirow{3}{*}{\rotatebox{90}{Label}} & Pass & 169 & 5 & 0 & 0.9713 & 0.9903 \\ \cline{2-7} 
				& Ques & 2 & 145 & 0 & 0.9864 & 0.9831 \\ \cline{2-7} 
				& Fail & 0 & 0 & 59 & 1.0000 & 1.0000 \\ \hline
		\end{tabular}}\label{tab:T2-noisy}
	\label{tab:noisy}
\end{table}

\section{Discussion and Conclusion}\label{sec:Conclusions}
In this paper, we have proposed a deep learning method for IQA of pediatric T1- and T2-weighted MR images. 
The network consists of a nonlocal residual network (NR-Net) for slice IQA and a random forest for volume IQA. 
Our method requires only a small amount of quality-annotated images for pre-training the NR-Net for initial automated annotation of a large amount of unlabeled dataset. This is refined via subsequent self-training processes.
Our method copes with label noise effectively, affords tolerance to inevitable rater errors, minimizes the amount of required label data, and reduces the number of manual labeling hours.
Experimental results verify that the proposed method yields near perfect IQA accuracy at a very low computational cost.

The proposed two-stage method is flexible and can be adjusted according to data availability. 
When only slice labels are available, NR-Net can be trained for slice IQA. 
When only volume labels are available, the two-stage framework can be used to identify slices that are useful for network training for slice and volume IQA. 
When a large labeled dataset is available, the semi-supervised learning process can be omitted. 
When a small labeled dataset and a large unlabeled dataset are available, the whole proposed framework can be utilized. 

The proposed method can be combined with quality enhancement methods, such as artifact removal methods, for integrated quality control.
The proposed framework can also be adapted to other IQA tasks, for instance, the IQA of diffusion MRI, where the size of unlabeled data is typically larger than that of labeled data, and where label noise is a significant issue. 
Future efforts will be directed towards such extension of our method.

\section*{Acknowledgement}
This work was supported in part by NIH grants (AG053867, EB006733, MH117943, MH104324, and MH110274) and the efforts of the UNC/UMN Baby Connectome Project Consortium.

\footnotesize
\bibliographystyle{IEEEtran}
\bibliography{BibLibrary}

\end{document}